# Process Makna - A Semantic Wiki for Scientific Workflows

Adrian Paschke, Zhili Zhao

Freie Universität Berlin
[Paschke|Zhili] AT inf.fu-berlin.de

**Abstract:** Virtual e-Science infrastructures supporting Web-based scientific workflows are an example for knowledge-intensive collaborative and weakly-structured processes where the interaction with the human scientists during process execution plays a central role. In this paper we propose the lightweight dynamic user-friendly interaction with humans during execution of scientific workflows via the low-barrier approach of Semantic Wikis as an intuitive interface for non-technical scientists. Our Process Makna Semantic Wiki system is a novel combination of an business process management system adapted for scientific workflows with a Corporate Semantic Web Wiki user interface supporting knowledge intensive human interaction tasks during scientific workflow execution.

## 1. Introduction

Scientific Workflows underlie the large-scale complex e-science applications and infrastructures, e.g., lab experiments, simulations, complex knowledge-intensive research questions. Compared with business workflows, a scientific workflow has special features such as computation, data or transaction intensity, knowledge-intensive human expert interaction, and a large number of activities including management of virtual organisations such as research collaborations and research teams. The described emerging computing infrastructures with powerful computing and resource sharing capabilities present the potential for accommodating those special features.

There are many industrial-strength Business Process Management (BPM) workflow tools available. However when it comes to knowledge-intensive scientific workflows, practical experiences have shown that the modelled scientific process representations are often not enacted correctly, consistently and homogenously. The reality of daily work of a researcher involves different science domains, different roles, and heterogeneous information and data objects. Moreover, the scientific processes are subject to frequent changes and exceptions which require flexible compensations and many human interactions. Traditional BPM approaches such as "Peoplelinks", "Partnerlinks",

"Exceptions", "Compensations" provide limited support for coordination, collaboration and integration in scientific workflows due to the lack of semantics and intuitive easy to use interfaces for human computer interactions.

In this paper, we present a solution for this through the integration of our Process Makna Semantic Wiki with a BPM workflow system. While the Wiki makes it easy to collaboratively author and add the required input data at runtime at the user front-end, the BPM system running in the back-end executes and enforces the process workflow. The Semantic Web extension of the Wiki allows establishing an explicit domain model in the back-end which represents the knowledge about the research domain such as ontologies and tasks / goals / as declarative rules and reaction rules. This additional semantic knowledge base is used to integrate heterogeneous information resources and services (i.e. semantic web services), mediate between different domains and provide semi-automated decision support and enforcement via rule reasoning and execution. Ultimately this leads to a novel eScience Wiki environment - called Process Makna (Makna stands for "knowledge" in Indonesian) - which is in particular tailored for knowledge-intensive agile scientific processes, typically realized through more weakly structured workflows with many human interactions.

The rest of the paper is organized as follows: We first present an example of a scientific workflow to illustrate the improvement coordination, collaboration and integration support by integration of Process Makna with BPM system. In Section 3 we propose the model integrating the semantic Wiki's data model with a BPM model and rule-based choreography style execution model. In Section 4 and 5 we discuss related work and the contribution of this work.

## 2. An e-Science Use Case

Virtual e-Science infrastructures supporting network-based scientific workflows are knowledge-intensive collaborative and weakly-structured processes. These virtual environments need to integrate the necessary heterogeneous scientific resources, services, data and documents. They need to interlink the information objects and transfer them into meaningful knowledge. And, they need to interact with the distributed human scientists to support them in their collaborative research workflows. In this section we present an example scientific workflow to illustrate improvement of coordination, collaboration and integration support with Process Makna as part of a virtual rule-based e-Science infrastructure, called Rule Responder. [Paschke08].

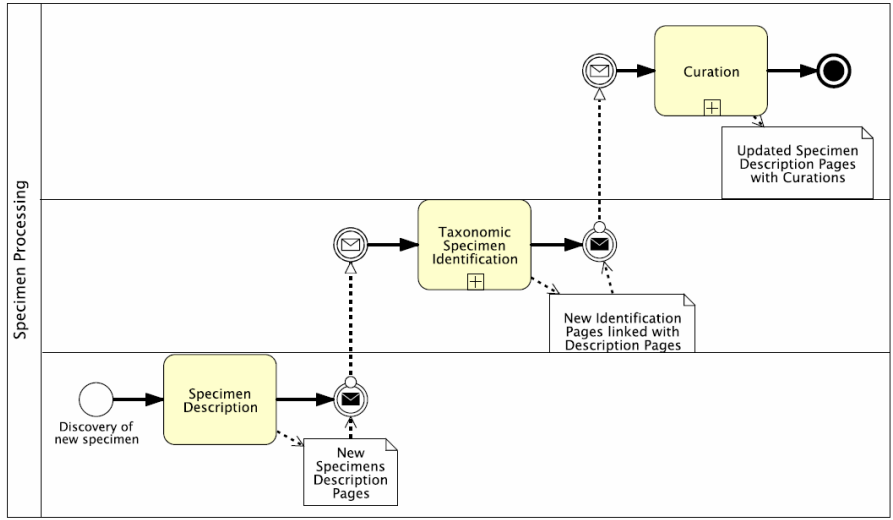

Figure 1: Sample workflow for the processing of a new specimen. The three rows are associated with the process swimlanes *FieldWorkParticipant*, *Taxonomist* und *Curator*.

Figure 1 shows a typical workflow: the processing and description of a specimen. It involves three actors: a field trip participant who describes the finding of the specimen, a taxonomist who performs the taxonomical identification and a curator who is responsible for the final curation of the descriptions. First, somebody in the role of a *field work participant* triggers the start of the process[1] and describes the discovery of a new specimen through an associated task form in the user console, whereby a new Wiki page is created from a template. This template contains enhanced Wiki syntax with placeholders for subjects of statements, which are replaced with corresponding user input. User input consists of literals (e.g. from text area and input fields) as well as URIs of concepts and instances. Predicates used in the template are taken from existing ontologies.[2] Notification by email is sent to somebody in the *taxonomist* swimlane to inform him/her that he/she is due to perform a taxonomic identification of this specimen. Again, the results are committed to the workflow engine through a Wiki task form. Upon completion a new page with the results of the taxonomic identification is created, and a typed link to it is inserted on the discovery page. A last action must be taken by somebody in the role of a *curator*.

The automated expert selection is based on semantic description of curators' responsibilities and the results of the taxonomist's identification. and user resources. Further on, Process Makna's click-searches which are provided with every resource in the Wiki can also be helpful for collaboration. Examples include a list of all users with tasks in a certain process and a list of all specimen that have been identified by a certain taxonomist.

---

[1] Workflow related functionality (e.g. listing and starting of processes) is provided by JSPWiki plugins in Makna which use RMI to interact with the jBPM engine.

[2] In our example we reuse the FungalWeb ontology [SNBHB05] for mycological classifications and TDWG's LSID ontologies for taxonomic data.

Another aspect of collaboration support is the selection of an actor in the curator swimlane based on the results of the taxonomic identification and the semantic responsibilities descriptions in the Wiki. In this example the taxonomic identification requires the selection of a *FungalWeb* concept and responsibility descriptions factors refer to the same ontology. Because of the hierarchical structure of the *FungalWeb* ontology it is possible to infer the responsible actor, thus realizing a dynamic assignment strategy.

The process of the specimen identificiation in the classification subprocess is modelled in the following figure 2.

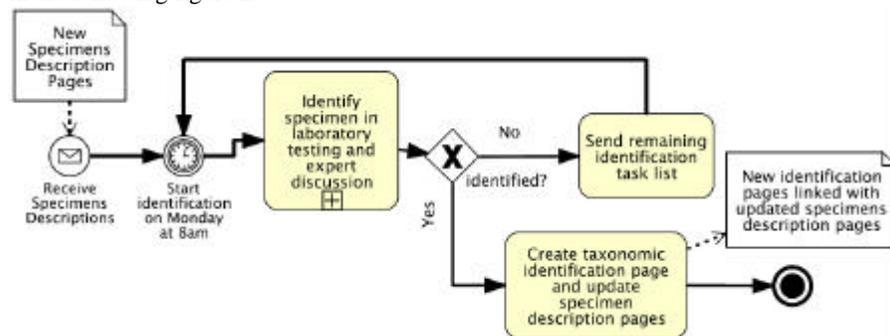

Figure 2: Abstract Model of an Experimental Study

The discussion about the final curation takes place in the Wiki, and in schedules telephone conference calls as modelled in figure 3.

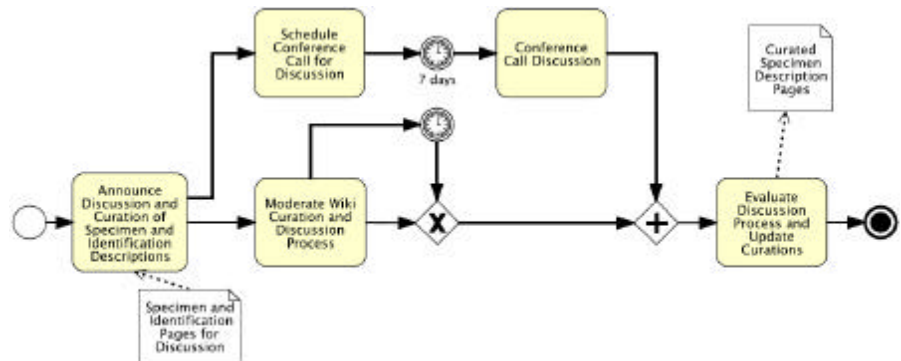

Figure 3: Coordination / Discussion Process between the different roles in the Scientific Workflow

## 3. Rules for Choreography Scientific Workflow Execution

Process Makna is a Wiki-based tool for distributed knowledge engineering. It extends the existing Makna Wiki engine with generic, easy-to-use ontology-driven components for collaboratively authoring, querying and browsing Semantic Web information. The architecture of Makna consists of the Wiki engine JSPWiki (http://www.jspWiki.org/), extended with several components for the manipulation of semantic data, and the underlying persistent storage mechanisms. A more detailed description of core Makna Wiki and the user interfaces it supports can be found in [DPT06].

For this work we integrated the semantic Wiki Makna and the workflow engine jBPM *(jboss.com/products/jbpm)* and the RuleML Rule Responder ESB middleware *(responder.ruleml.org/)*. The architecture of the distributed system with jBPM and Makna consists of a J2EE server, J2SE server(s), and a workstation and AllegroGraph triple store server (agraph.franz.com/*)* that serve as persistence stores. In this section we consider the aspects of this integration which were implemented for Process Makna.

### 3.1 Semantic Workflow Annotation and Rule Integration

It is necessary to reflect the workflow instances (e.g. tasks and processes) in the semantic model in order to support search for and enhanced presentation of tasks and processes. The insertion of workflow concepts into the semantic model at runtime has two prerequisites: first, a mechanism must be provided which assigns the URIs of the ontology concepts to the scientific workflow models and the domain-specific workflow instances; second, these URIs must be made accessible in the process execution definitions. This integration of ontologies has the advantage of allowing access to existing domain specific glossaries, taxonomies and ontologies from within the processes. Different scientific vocabularies, e.g. from different research domains, can be mediated via the semantic inference capabilities. With these prerequisites met, semantic decoration of workflow execution can be performed with standard jBPM procedures. While traversing a process graph the engine fires events (event messages) – e.g. *process-start*, *process-end*, *task-start*, *task-end* and *task-assign* – which are associated with reaction rules which trigger custom actions. In our rule-based extension for Process Makna arbitrary complex scientific workflow logic can be declaratively represented in terms of combinations of derivation and reaction rules which associate the workflow events with conditions (implemented as derivation rule sets) and actions that perform e.g. RDF update queries based on URIs for workflow instances, i.e. the progress of a workflow can be reflected in the semantic model.

### 3.2 Semantic Flow Conditions

Facilitating semantically enhanced flow conditions in the process execution phase is desirable, because inference allows for transitions and control flow, based on the description of a workflow in the semantic model itself. jPDL and BPEL already support the association of transitions with boolean expressions (or simple rules in BPEL) which are based on workflow relevant data. Our rule-based extension provides adequate high expressiveness for complex declarative flow conditions. The process execution continues via the first transition whose associated expression resolves to true, or via the default transition if none of the expressions resolves to true, though this behavior can be customized in a rule-based way. This procedure facilitates simple rule-based control of the process based on semantic model state.

### 3.3 Semantic Assignments

Strategies for the assignment of tasks to Wiki users that build on semantic user and task descriptions are supported by the system. Semantic user descriptions can be based on customized formalizations such as DOAC (http://ramonantonio.net/doac/). Task descriptions and responsibility assignments are a bit more complicated because not only the general description of a task, which is valid for all task instances, but also details of the current execution of a particular instance might be relevant for assignments. The description of users and tasks can use common concepts, e.g. by sharing domain ontologies such as the responsibility assignment matrix (ruleresponder.ruleml.org). The actual assignment of tasks to users is realized through a jBPM *AssignmentHandler* implementation which can be configured with a SPARQL query that references these concepts.

### 3.4 Semantic Search and Presentation of Workflow Individuals

For the structured presentation of semantic resources we have added support for formatting SPARQL XML responses with XSLT. This functionality is encapsulated as another JSPWiki-Plugin, which additionally has support for expressions that are resolved at rendering time (e.g. logged in user and page URL). Via an endpoint parameter remote triplestores are also supported. The plugin enables the structured presentation of workflow individuals such as tasks and process instances in the Wiki. It can be called from the JSPWiki template level (invoked from a JSP) or from the Wiki page level (invoked from Wiki syntax). Another application of this plugin in the generation of lists such as semantically enhanced task lists. The tasks that have been assigned to the logged in user can be arranged by domain specific structures, thus enabling semantically enhanced task lists.

### 3.5 Rule-based Workflow Execution

Currently, languages for describing business processes and business interaction protocols in a machine-readable way such as the jPDL and BPEL only provide very simple qualifying conditions. Expressive rule sets which are able to represent complex conditional decisions, task/goals, and reactions (e.g. event condition action rules), as need in scientific workflows, are not supported.

We have implemented a rule based extension of jPDL and BPEL which exploits Reaction RuleML (ruleml.org and Reaction RuleML) as platform-independent Web rule standard and the Rule Responder ESB middleware (responder.ruleml.org) with the Prova (prova.ws) rule engine as platform-specific execution environment. The conditional control flow of a business process is defined by the order of sending and receiving message constructs in Prova's messaging reaction rules. These messaging reaction rules maintain a local conversation state which reflects the process execution state and support performing of different activities within process instances managed in simultaneous conversation branches. The behaviour of the inbound links defined by the messaging reaction rules correspond to the BPEL links, but provide an additional rule execution layer with a much more expressive and compact declarative rule-based programming language to represent complex event processing logic in arbitrary combination with conditional decision logic implemented in terms of derivation rules.

As an independent system, the Rule Responder Enterprise Service Bus middleware with the Prova rule engine is running as a parallel running platform that analyses and processes events. The BPM- and the RuleResponder-platform correspond via events which are produced by the jBPM-workflow engine and by the IT services which are associated with the business process steps.

**Rule Responder** (http://responder.ruleml.org) is a middleware for distributed intelligent rule-based inference services and complex event processing on the Web. It adopts the OMG model driven architecture (MDA) approach:

1. On the computational independent level rules are engineered computational independent e.g. in a natural controlled English language using blueprint templates and user-defined controlled vocabularies.
2. The rules are mapped and serialized in a interchange format such as Reaction RuleML (or RIF XML) which is used as platform independent rule interchange format to interchange rules between the rule inference services and other arbitrary execution environments.
3. The Reaction RuleML rules are translated into the platform specific rule language for execution, e.g. the Prova rule language.

Arbitrary (reaction) rule engines and CEP engines can be deployed on an enterprise service bus (ESB). Translator services translate from the respective execution syntax of the platform-specific rule / CEP engine into Reaction RuleML which is used as common rule interchange format and vice versa. Arbitrary transport protocols (>30 protocols) such as MS, SMTP, JDBC, TCP, HTTP, XMPP can be used to transport rule sets, queries and answers (e.g. event instance data) between distributed inference service endpoints on the Web. For workflow-like intelligent message processing it employs the Web rule engine Prova (http://prova.ws) and messaging reaction rules to describe (abstract) processes / situations and complex event processing logic in terms of message-driven conversations between inference services / agents, i.e. it represents their associated interactions via constructs for asynchronously sending and receiving event messages. The control flow of a process is defined by the order of sending and receiving message constructs in messaging reaction rules. In contrast to standard Event-Condition-

Action (ECA) rules which typically only have one global state, messaging reaction rules maintain a local conversation state which reflects the process execution state and support performing of different activities within process instances managed in simultaneous conversation branches.

## 4. Related Work

First scientific workflow management systems such as Gridbus workflow[3], SwinDeW-G[4], Kepler[5] and Taverna[6] are developed or evolved from existing systems. While these systems in general address the automation of eScience infrastructures, we specifically contribute in this paper with an approach of exploits the user-friendliness of a Wiki as regarding multi-site content generation and the power of semantic technologies in combination with BPM technologies as with respect to organizing and retrieving scientific knowledge to support weakly-structured processes which involve knowledge intensive human interactions and are subject to frequent changes and agile compensations and exceptions.

There are many Wiki systems. TWiki has been an early perl/cgi implementation [TWiki]. MediaWiki is a free Wiki implementation based on PHP. It was originally written for Wikipedia, but is broadly used today [MedWiki]. Semantic Media Wiki [SemWiki] is a well-known semantic Wiki implementation with the purpose to create a Semantic WikiPedia [SeWiPe]. IkeWiki is a semantic Wiki created by Salzburg Research [IkeWiki]. OntoWiki is another semantic Wiki system [OntoWiki]. To the best of our knowledge none of these (semantic) Wikis has an extension for extended BPM workflow models and is used as human-friendly interface during scientific workflow execution and enforcement.

Semantic Business Process Management (SBPM) and semantic workflows have been addressed by various projects in the past, e.g. the Semantic Business Process Management Working Group [SBPM], the Super project [Super], the Rule-based Service Level Agreement Project (RBSLA) [RBSLA], or the Semantic Business Process Management workshop series [SBPMW]. The contributions in these projects mainly address the combination of ontologies and/or rules with business process models, and the enriching of web services to semantic web services including non-functional properties and policies / SLAs. While these approaches have demonstrate the usefulness of Semantic Web technologies for BPM and automated IT service execution within business processes they have not addressed the problems of knowledge-intensive weakly structured processes with many human interactions which motivated our work in this paper. Nevertheless our work is fundamentally based on the findings of these earlier projects.

---

[3] http://www.gridbus.org/workflow/
[4] http://www.swinflow.org/swindew/grid/
[5] https://kepler-project.org/
[6] http://taverna.sourceforge.net/

The current extensions of BPM languages such as BPEL4People [BPEL4P] are purely syntactic and introduce "Peoplelinks", "Partnerlinks", "Exceptions", "Compensations", "HumanTasks" etc. They do not address the semantics of these interactions and do not investigate the combination of social software + semantic technology (=Web 3.0) for scientific workflows.

## 5. Conclusion

Manifold rigorous and industry-strength BPM languages [7] and tools have been developed. Typically these tools are applied in repetitive, high-volume production or administrative workflows where a central orchestration process definition can be created, optimized, and then used as a prescriptive, standardized model for process enactors or for workflow automation. However, practical experiences have shown that scientific workflows process are often weakly-structured, agile / change quickly, and need to dynamically integrate heterogeneous information resources and services as well as collaborative human knowledge and tasks.

We have considered in this paper how the limited coordination, collaboration and (human) integration support of BPM approaches could be improved for scientific workflows by a Web 3.0 semantic process Wikis and have presented a solution based on integration with a BPM workflow system. We see two major advantages of our proposal:

1. decentral, lightweight dynamic acquisition of required process knowledge – (human) end users bring in their process knowledge in a simple, effortless, manner - the low-barrier Wiki approach provides and intuitive interface to collect knowledge from humans.
2. the ongoing collaborative (conversation-based) human interaction may be very simple, yet powerful, instrument to execute rule-based agile scientific workflows processes.

Initial evaluation of our implementation in a HCLS e-Science infrastructure [Paschke2008] has demonstrated promising improvements in these aspects. Guided by the scenario, described in section 3, several tests were run to ensure that the prototype mainly performs correctly and stable in the expected behavior. In particular, a set of defined use cases covering the main functionalities have been performed. Additionally, during the development of the prototype several unit tests have been run to check the correct implementation of the various methods. Thus, it can be stated that the prototype generally performs correctly and in the authors' view, the evaluated use cases can prove the importance of the future uptake of semantic Wiki systems as user-friendly interface for human interaction during scientific workflow execution, particularly in corporate environments.

---

[7] Currently, it seems that BPMN (for modelling) and BPEL (for execution) will become the broadly accepted standards

Beside the contribution of a new design artefact – a semantic BPM Wiki for Scientific Workflows – the extension of the existing standards BPMN and business process executionwith Semantic Web ontologies and rules provides significant more expressiveness and leads to a novel semantic BPM (SBPM) approach. Moreover, a typical problem of Corporate Semantic Web Wikis is that they are rarely moderated and hence quickly grow with a lot of redundant, outdated or simply wrong information. Current and obsolete information cannot be distinguished, and it is hard to find the relevant information and navigate the Wiki. A process model for content authoring which runs in the back-end of the Wiki can serve this role of a moderator and can enforce respective content management policies.

We are continuing the development of Process Makna, and particularly currently explore in a practical manner the deployability of it in different enterprise environments through our Corporate Semantic Web project (http://www.corporate-semantic-web.de), which has as one of its goals research in corporate semantic collaboration. In particular, we are currently implementing a larger virtual e-Science infrastructure within the scope of the W3C HCLS interest group.

### Acknowledgments

This work has been partially supported by the "InnoProfile-Corporate Semantic Web" project funded by the German Federal Ministry of Education and Research (BMBF) and the BMBF Innovation Initiative for the New German Laender - Unternehmen Region.### References

[DPT06] Karsten Dello, Elena Bontas Simperl Paslaru, and Robert Tolksdorf. Creating and
using semantic web information with makna. In Max V¨olkel and Sebastian Schaffert, editors, *Proceedings of the First Workshop on Semantic Wikis - From Wiki To Semantics*, volume 206 of *Workshop on Semantic Wikis*, pages S.43–57, Budva, Montenegro, June 2006. ESWC2006.

[SNBHB05] Arash Shaban-Nejad, Christopher Baker, Volker Haarslev, and Greg Butler. The
fungalweb ontology: Semantic web challenges in bioinformatics and genomics. In *Proceedings of the International Semantic Web Conference 2005*, volume 3729 of *Lecture Notes in Computer Science*, pages 1063–1066. Springer, 2005.

[Paschke08] Paschke, A. 2008. Rule responder HCLS eScience infrastructure. In *Proceedings of the 3rd international Conference on the Pragmatic Web: innovating the interactive Society* (Uppsala, Sweden, November 28 - 30, 2008). P. J. Ågerfalk, H. Delugach, and M. Lind, Eds. ICPW '08, vol. 363. ACM, New York, NY, 59-67. DOI= http://doi.acm.org/10.1145/1479190.1479199

[TWiki] TWiki, http://tWiki.org